\pdfoutput=1

\documentclass[11pt]{article}

\usepackage[table]{xcolor}
\usepackage{tikz}
\usetikzlibrary{tikzmark}

\usepackage{emnlp2021}

\usepackage{times}
\usepackage{latexsym}

\usepackage[T1]{fontenc}

\usepackage[utf8]{inputenc}

\usepackage{booktabs}
\usepackage{multirow}
\usepackage{colortbl}

\usepackage{graphicx} 

\usepackage{soul}
\usepackage{enumitem}

\newcommand{\presup}[1]{\textit{#1}}
\newcommand{\trigger}[1]{{#1}}
\newcommand{\nli}[1]{\textsc{#1}}

\usepackage{pgf}
\usepackage{collcell}

\def\colorModel{hsb}
\newcommand\ColCell[1]{
  \pgfmathparse{#1<30?1:0}  
    \ifnum\pgfmathresult=0\relax\color{white}\fi
    \pgfmathsetmacro\compA{210/360} 
    \pgfmathsetmacro\compB{(#1)/30}       
    \pgfmathsetmacro\compC{0.9}            
  \edef\x{\noexpand\centering\noexpand\cellcolor[\colorModel]{\compA,\compB,\compC}}\x #1
  } 
\newcolumntype{E}{>{\collectcell\ColCell}m{4.5ex}<{\endcollectcell}}  

\def\colorModel{hsb}
\newcommand\ColCellx[1]{
  \pgfmathparse{#1<30?1:0}  
    \ifnum\pgfmathresult=0\relax\color{white}\fi
    \pgfmathsetmacro\compA{210/360} 
    \pgfmathsetmacro\compB{(#1)/80}       
    \pgfmathsetmacro\compC{0.9}            
  \edef\x{\noexpand\centering\noexpand\cellcolor[\colorModel]{\compA,\compB,\compC}}\x #1
  } 
\newcolumntype{F}{>{\collectcell\ColCellx}m{4.5ex}<{\endcollectcell}}  

\usepackage{microtype}

\usepackage{linguex}

\title{NOPE: A Corpus of Naturally-Occurring Presuppositions in English}

\author{
Alicia Parrish${}^*$ $\quad$ Sebastian Schuster${}^*$ $\quad$ Alex Warstadt\thanks{\ \ \ Equal contribution.}\\
\textbf{Omar Agha  \hskip0.75em
Soo-Hwan Lee \hskip0.75em 
Zhuoye Zhao \hskip0.75em}  
\textbf{Samuel R. Bowman \hskip0.75em
Tal Linzen} \\
Department of Linguistics \& Center for Data Science, New York University \\
\texttt{\{alicia.v.parrish,schuster,warstadt\}@nyu.edu}}

\date{}

\begin{document}
\maketitle
\begin{abstract}
Understanding language requires grasping not only the overtly stated content, but also making inferences about things that were left unsaid. 
These inferences include \textit{presuppositions}, a phenomenon by which a listener learns about new information through reasoning about what a speaker takes as given. Presuppositions require complex understanding of the lexical and syntactic properties that trigger them as well as the broader conversational context.
In this work, we introduce the Naturally-Occurring Presuppositions in English (NOPE) Corpus to investigate the context-sensitivity of 10 different types of presupposition triggers and to evaluate machine learning models' ability to predict human inferences. We find that most of the triggers we investigate exhibit moderate variability. We further find that transformer-based models draw correct inferences in simple cases involving presuppositions, but they fail to capture the minority of exceptional cases in which human judgments reveal complex interactions between context and triggers.
\end{abstract}

\section{Introduction} 
\label{sec:intro}

In every statement, certain facts are taken for granted by the speaker; such facts, while left unsaid, can often be inferred by the listener. A speaker who says \emph{Chet finished law school}, for example, asserts that Chet reached the end of law school, and \emph{presupposes} that he attended law school in the first place. While such presuppositions are usually not the main information that a speaker intends to convey, listeners can still learn new information from them by drawing inferences about what the speaker takes as a given. 

\begin{figure}[t]
    \centering
    \includegraphics[width=\columnwidth]{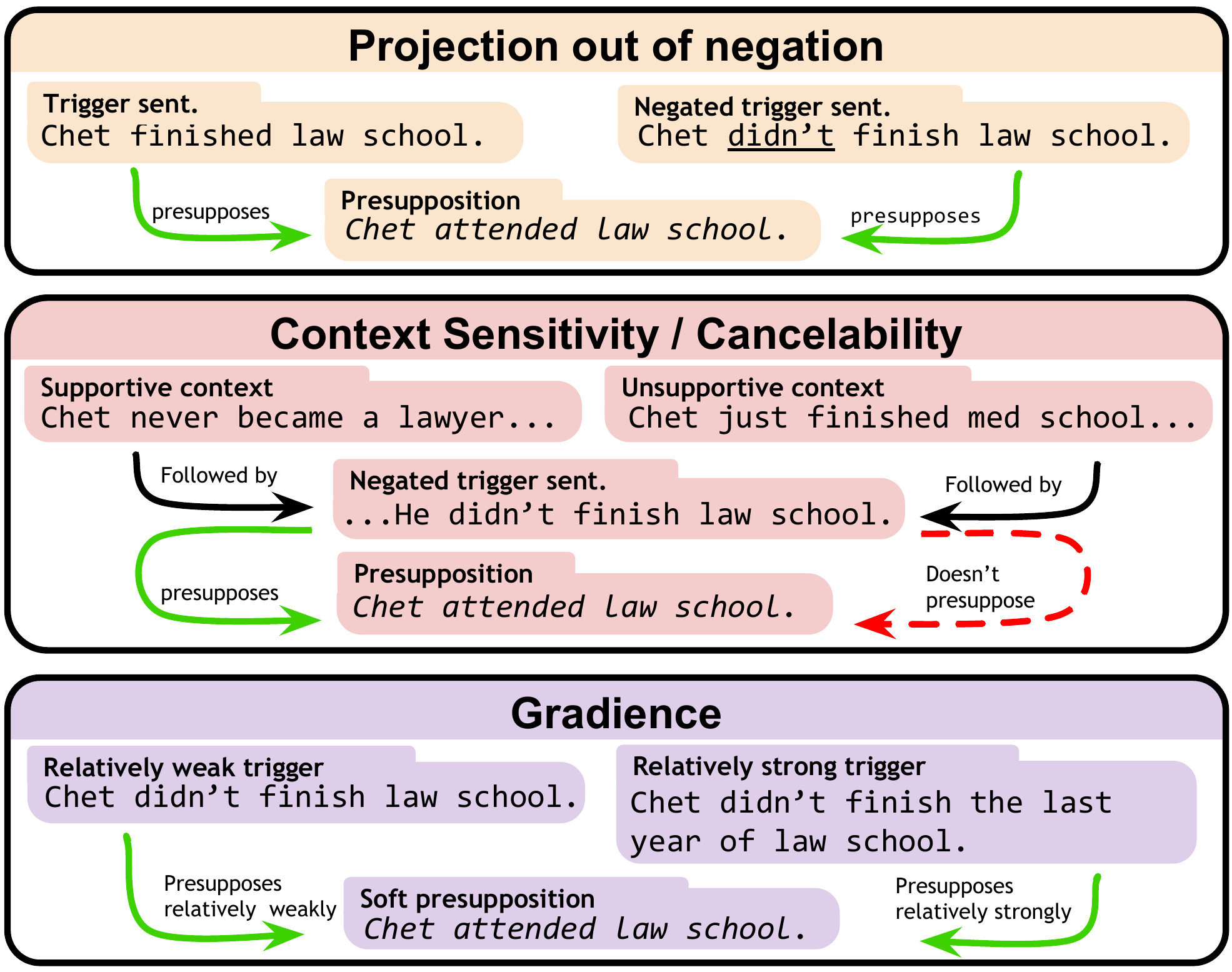}
    \caption{Inferences involving presuppositions can be challenging because presuppositions project out of negation (first box), they are context-sensitive and can be canceled (second box), and they can differ in how easily they can be cancelled and how likely a listener is to make the inference (third box).}
    \label{fig:1}
\end{figure}

One type of signal that helps human listeners draw such inferences is presupposition \emph{triggers}, such as the verb \emph{finish}. Triggers have long been known in linguistics to be associated with presuppositions, but the nature of this relationship is a matter of much debate \cite[e.g.,][]{heim1983projection,VanderSandt1992,chemla2009similarity,abusch2010presupposition,simons2010projects,romoli2015presuppositions,abrusan2011predicting,schlenker2021triggering}. This is in part because presuppositions and triggers exhibit properties (summarized in Figure \ref{fig:1}) that make the resulting inferences heterogeneous and make precise theory development challenging. How humans draw inferences involving a wide variety of triggers is therefore still poorly understood, and this lack of a comprehensive theory makes it difficult to construct test cases for machine learning models with as much diversity as found in naturalistic data.

We introduce a broad-coverage dataset of sentences with naturally-occurring presupposition triggers, called NOPE (\textbf{N}aturally-\textbf{O}ccurring \textbf{P}resuppositions in \textbf{E}nglish). We show that this corpus can be used to analyze the variability associated with different presupposition triggers, and we use this dataset to investigate how well current natural language understanding models \cite[specifically natural language inference, or NLI, models; ][]{Dagan2006} are able to make similar inferences.

We find considerable variability in human ratings across items for two types of triggers, clause-embedding predicates (e.g, \textit{know} and \textit{think}) and implicatives (e.g., \textit{manage to} and \textit{fail to}), in line with previous work \cite{demarneffe2019commitmentbank,ross2019well}, and we find that the other triggers exhibit lower contextual variability. Further, transformer-based models fine-tuned on NLI are largely able to draw correct inferences for simple cases involving presupposition triggers but they fail to fully capture the contextual variability and gradience in the human judgments. We release our dataset and code at \url{https://github.com/nyu-mll/nope}.

\section{Background}

\paragraph{Properties of presuppositions} One important property of presuppositions is that they \textit{project} out of environments like negation \cite{karttunen1973presuppositions,heim1983projection}. While negating a sentence cancels its entailments, as the top box of Figure \ref{fig:1} illustrates, both the ``trigger sentence'' and its negation give rise to the same presupposition. Second, presuppositions, unlike entailments, can be canceled and disappear in certain contexts \cite{simons2001conversational,abusch2002lexical,simons2010projects}. 

Third, presupposition triggers exhibit gradience both in the sense that a listener may be more or less confident about inferring what a speaker presupposes upon hearing different triggers \cite{tonhauser2018how,tonhauser2020which,degen2021prior,Mahler2020}, and in the sense that the presuppositions of different triggers may be more or less easy to cancel \cite{abusch2002lexical,demarneffe2019commitmentbank}. Specifically, some triggers are considered \textit{hard} triggers, which require the presupposition to be satisfied for the statement to be well-formed. For example, \textit{There are three apples and both of them are green} is not well-formed because it involves \textit{both}, a hard trigger that always presupposes that there are two objects. \textit{Soft} triggers such as the change of state verb \textit{finished}, on the other hand, can easily be cancelled, as illustrated in the second box in Figure~\ref{fig:1}.

\paragraph{Existing resources} Recent work in NLP has investigated the ability of neural networks trained for natural language understanding to pick up on subtle discourse cues \cite[e.g,][]{upadhye2020predicting,schuster2020harnessing}. Among them, some have introduced resources specifically targeting presuppositions \cite{white-etal-2017-inference,demarneffe2019commitmentbank,jeretic2020impppressive} and tested neural models on them \cite{white2018lexicosyntactic,jiang2019evaluating,jeretic2020impppressive,ross2019well}. However, this previous work has either focused on a specific class of presupposition triggers (e.g., clause-embedding verbs or implicatives), or has made the simplifying assumption that context does not affect presuppositions and therefore did not include naturally-occurring contexts. Our NOPE corpus complements existing corpora by including a wider range of trigger types. 

\newcite{Kim2021} recently also demonstrated the importance of detecting and verifying presuppositions for natural language understanding tasks such as question answering. They found that verifying presupposed content in questions results in question-answering systems that generate more helpful responses to unanswerable questions.

\section{Dataset Construction}

\begin{table*}[ht!]\small\centering
    \rowcolors{1}{white}{gray!25}
    \begin{tabular}{p{0.12\linewidth}p{0.27\linewidth}p{0.27\linewidth}p{0.25\linewidth}}
    \toprule
    Trigger &  Affirmative Example & Negative Example & Presupposition\\
    \midrule
    Change of state & A microsecond later, images from his exterior sensors \ul{snapped} into focus. & A microsecond later, images from his exterior sensors \textbf{didn't} snap into focus. & Previously, images from his exterior sensors hadn't been in focus.\\
    Clefts & But \ul{it is} the horse racing \ul{that} is just for children. & But it \textbf{isn't} the horse racing that is just for children. & There's something that is just for children \\
    Comparatives & That is \ul{a bigger problem, than} the chairman's claim. &  That \textbf{isn't} a bigger problem, than the chairman's claim. & The chairman's claim is a problem. \\
    Aspectual verbs & At the age of 55, I \ul{began} preparing myself to die. & At the age of 55, I \textbf{didn't} begin preparing myself to die. & Before age 55, I was not yet preparing to die. \\
    Embedded questions & I fail to \ul{see how} you can rationalize rewarding illegality. & I \textbf{don't} fail to see how you can rationalize rewarding illegality. & You can rationalize rewarding illegality. \\
    Clause-embed. verbs & In 20 years we'll \ul{realize} that's a mistake. & In 20 years we \textbf{won't} realize that's a mistake. & [Pushing people towards pharmaceuticals] is a mistake. \\
    Implicatives & The survivors \ul{managed} to scramble out through the tiny gap in the rocks. & The survivors \textbf{didn't} manage to scramble out through the tiny gap in the rocks. & The survivors made an attempt to scramble out through the tiny gap in the rocks. \\
    Numeric determiners & \ul{Both} protagonists in the room defy a political force and receive aid from a higher authority. & Both protagonists in the room \textbf{do not} defy a political force and receive aid from a higher authority. & There are two protagonists in the room.\\
    ``Re-'' prefixed verbs & Taoism \ul{reconnects} aging to the great cycles of nature. & Taoism \textbf{doesn't} reconnect aging to the great cycles of nature. & Aging was once connected to the great cycles of nature. \\
    Temporal adverbs & He took them to the NL Championship Series last year \ul{before} being swept by the Atlanta Braves. & He \textbf{didn't} take them to the NL Championship Series last year before being swept by the Atlanta Braves. & Johnson was swept by the Atlanta Braves.\\
    \bottomrule
    \end{tabular}
    \caption{Selected examples present in NOPE. Presupposition triggers are \ul{underlined}.} \label{tab:all triggers}
\end{table*}

\subsection{Trigger selection}

We identify 10 presupposition trigger types to focus on based mainly on \citeauthor{levinson1983pragmatics}'s (\citeyear{levinson1983pragmatics}) widely taught list \cite[see][for other similar lists]{beaver1997presupposition,potts2015presupposition}. These specific triggers were selected because they are common in English and systematic enough that we can extract them from a corpus. Table~\ref{tab:all triggers} contains a list of all trigger types in the present study, along with an example from the full dataset. Appendix~\ref{sec:trigger_types} includes a more detailed discussion of each trigger type and the presuppositions they generally give rise to. In this work, we focus mainly on soft triggers since we expect them to exhibit more context-sensitivity. As a control for the judgment paradigm, we also include the hard triggers \trigger{clefts} and \trigger{numeric determiners}, for which we expect humans to endorse the purported presupposition highly and consistently.

\subsection{Extraction of Examples}\label{3.1}  
We extracted sentences with presupposition triggers from the Corpus of Contemporary American English \citep[COCA, ][]{davies2008corpus}. Following \citet{demarneffe2019commitmentbank}, in addition to the trigger-containing sentence, we also extracted the two immediately preceding sentences for context and, where applicable, the speaker of each sentence. Because some spans of text in COCA are redacted, we only extracted contexts that form a contiguous span with the trigger-containing sentence. We also detokenized text and removed HTML tags.

We identified trigger-containing sentences in two stages. We first automatically extracted sentences using syntactic features extracted from SpaCy dependency parses \cite{spacy} or using lists of lexical items. For example, sentences with \trigger{clefts} such as \textit{It is the president who has to sign the document} can be identified by their syntactic structure whereas open-class lexical triggers such as \trigger{change of state} predicates (e.g., \textit{melt}) can only be identified using word lists (see Appendix~\ref{sec:word-lists} for the word lists used). 
Second, six of the authors, all of whom have graduate level training in formal linguistics, manually reviewed and annotated the extracted passages. All expert annotations were double checked by a native English speaker. We ensured both that the passage contains a genuine example of the trigger\footnote{For example, we excluded from the dataset sentences with a referential \textit{it} as in \textit{It is a multi-purpose bread that should be part of any culinary repertoire}, which resemble clefts, and were occasionally labeled as such by the extraction pipeline.} and that the trigger sentence could be negated.\footnote{In some cases, negating the clause with the presupposition trigger led to contradictions or pragmatically odd sentences. For example, \textit{The dog started digging and uncovered the bone} contains the trigger \textit{start} and negating \textit{start} would lead to the non-sensical sentence \textit{The dog didn't start digging and uncovered the bone.}} 
We also excluded examples with conditionals and examples in which the presupposition trigger was embedded under another predicate, since these examples could not be straightforwardly negated.

In some cases (12.9\% of examples), we wrote an altered version of the trigger sentence by making small edits in order to make the subsequent annotations possible. For example, many sentences can be negated after removing an adverb: While the original \textit{I know what Elissa's hobby is already} becomes much less natural when negated due to the presence of \textit{already}, the slightly altered sentence without the adverb, \textit{I know what Elissa's hobby is}, can easily be negated. We further systematically altered sentences for two types of triggers. For temporal adverbs, we rewrote sentences where the adverbial clause is preposed, in order to allow negation to scope over the adverbial (e.g. \emph{Before it rained, we danced} becomes \emph{We danced before it rained}). For numeric determiners, we added explicit domain restrictions in order to reduce vagueness in the spelled-out presupposition (e.g. \emph{Both cats meowed} becomes \emph{Both cats on the mat meowed}, with presupposition \emph{There are two cats on the mat}).

In total, we extracted 2,482 passages, of which 1,279 (51.5\%) met our criteria for inclusion. This resulted in more than 100 examples per trigger type for subsequent ratings.

\paragraph{Flipping the polarity of the trigger sentence}  
To investigate to what extent the presupposition projects out of negation, we manually constructed a negated version of the trigger sentence for each example. We added sentential negation (e.g. \emph{not}) to the main clause of the sentence or, in the case of clausal coordination, to the conjunct that contains the trigger. For the 102 corpus-extracted examples that already contained sentential negation, we removed the negation to create a non-negated version of the sentence.

\paragraph{Writing presuppositions} To test to what extent humans and NLI models infer the content of the presupposition for a given trigger sentence, it is necessary to spell out the content of the purported presupposition. We therefore wrote a sentence expressing the presupposition for each pair of non-negated and negated trigger sentences. For instance, for sentences with \trigger{change of state} predicates and \trigger{aspectual verbs}, we wrote sentences that refer to the state before the event described by the trigger sentence (e.g., \emph{Bill dropped the vase} presupposes \emph{Bill had been holding the vase just before then}). See Appendix~\ref{sec:writing-guidelines} for a description of our writing strategies for each trigger type. All sentences were checked by a second annotator and corrected if the second annotator discovered issues with the original phrasing.

\subsection{Adversarial Examples}  

A potential bias in the dataset is that the majority of examples give rise to presuppositions 
and therefore both the non-negated and the negated trigger sentences entail the sentence spelling out the presupposition most of the time. This creates an issue for model evaluation, since high accuracy in predicting the entailment relation between the trigger sentences and the presupposition sentence may be caused by some heuristic that leads to frequent predictions of \textsc{entailment}. This issue is further exacerbated by the fact that for many trigger types (e.g. \trigger{clause-embedding verbs}), the presupposition sentence tends to have very high lexical overlap with the trigger sentence and therefore a model that uses a heuristic to predict \textsc{entailment} when lexical overlap between the trigger sentence and the presupposition sentence is high (which has been found to be the case for many models, e.g., \citeauthor{mccoy2019right}, \citeyear{mccoy2019right}), will likely achieve high accuracy despite not being able to draw the correct inferences.

To control for this issue, we randomly selected 200 examples (20 per trigger type) from the main corpus and constructed adversarial sentences that differ minimally from the presupposition sentences but are no longer entailed.\footnote{For example, for the trigger sentence \textit{Women from both sides of town formed a mothers group}, we wrote the presupposition sentence \textit{There are two sides of town}, which is entailed by the trigger sentence. In the adversarial corpus, we changed the presupposition sentence to \textit{There are \textbf{three} sides of town}, which is no longer entailed by the trigger sentence.} 
This adversarial dataset in combination with the main dataset thus constitutes a set of minimal pairs as advocated for by \citet{gardner2020evaluating} and makes it possible to tease apart whether high accuracy on the main dataset is a result of a lexical overlap heuristic or more sophisticated inferences.

\subsection{Crowdsourced Probability Judgments}

We determined whether and how strongly naive participants infer the content of the expert-written presupposition after reading the passage containing the trigger for each example by crowdsourcing probability ratings. 
These ratings allow us to evaluate a) for which triggers the content of the purported presupposition projects out of negation, b) the context-sensitivity of a certain trigger type, and c) the level of gradience in inferences for different trigger types. Further, as we demonstrate in Section~\ref{sec:models}, we can use the ratings to evaluate how well NLI models mimic human inferences.

\paragraph{Task description}
For each example, we collected 5 probability judgments from participants via Amazon Mechanical Turk (MTurk). Each task consisted of 20-30 pairs of passages with presupposition triggers (the \textit{premise}) and the sentence capturing the content of the purported presupposition (the \textit{hypothesis}), presented one-by-one in randomized order. We further included 5 filler premise-hypothesis pairs from MNLI \cite{williams2020mnli} as attention checks (see Appendix~\ref{sec:crowdsourcing-appendix} for exclusion criteria). For each item, participants were presented with the following instructions: ``Indicate how likely you think the statement is to be true, using the information in the text and your background knowledge about how the world works.'' Participants provided ratings from 0.0 to 100.0 by adjusting a non-linear slider that allowed greater precision at the slider's edges, capturing that the distinction between 99.0 and 99.5\% probability is more informative than the distinction between 60.0 and 60.5\% probability \cite{tversky1981framing}. Slider endpoints were labeled ``impossible'' (0) and ``certain'' (100).

We used a pre-screening task to find reliable participants (see Appendix~\ref{sec:crowdsourcing-appendix}). Each task was advertised to qualified MTurk workers as taking 10 minutes to complete and paid USD 2.50 (USD 15/hr). Median completion time across all posted tasks was 9.6 minutes. Individual participants could provide a rating for at most 15\% of the total dataset.

\section{Analysis of Human Judgments}

To investigate to what extent the different trigger types give rise to presuppositions, and to what extent they exhibit context-sensitivity and gradience, we analyzed the crowdsourced human judgements as described in the following sections. 

\subsection{Mapping to NLI Labels}

To facilitate our analyses (and to allow us to use the widely-used NLI paradigm for evaluating models), we mapped the continuous ratings for each example to the three NLI labels \textsc{contradiction}, \textsc{neutral}, \textsc{entailment} by inferring upper and lower thresholds for \textsc{contradiction} and \textsc{entailment}, respectively. We determined these thresholds based on the participants' ratings of the filler items for which we have expected categorical judgments. Since different participants may use the scale differently, we computed individual thresholds for each participant (average for \textsc{contradiction}: 5.4\%, average  for \textsc{entailment}: 93.5\%; see Appendix~\ref{sec:nli-label-mapping} for details on the threshold inference procedure). We then assigned a label to each example based on the majority vote of the five participants. Following standard practice for NLI datasets \citep[e.g.,][]{bowman2015snli}, we discarded any examples without a majority label (see Table~\ref{tab:accuracy_comparison}). 

\subsection{Human Judgment Results}

\begin{figure}
    \centering
    \includegraphics[trim=0 0.4cm 0 0,clip,width=0.9\linewidth]{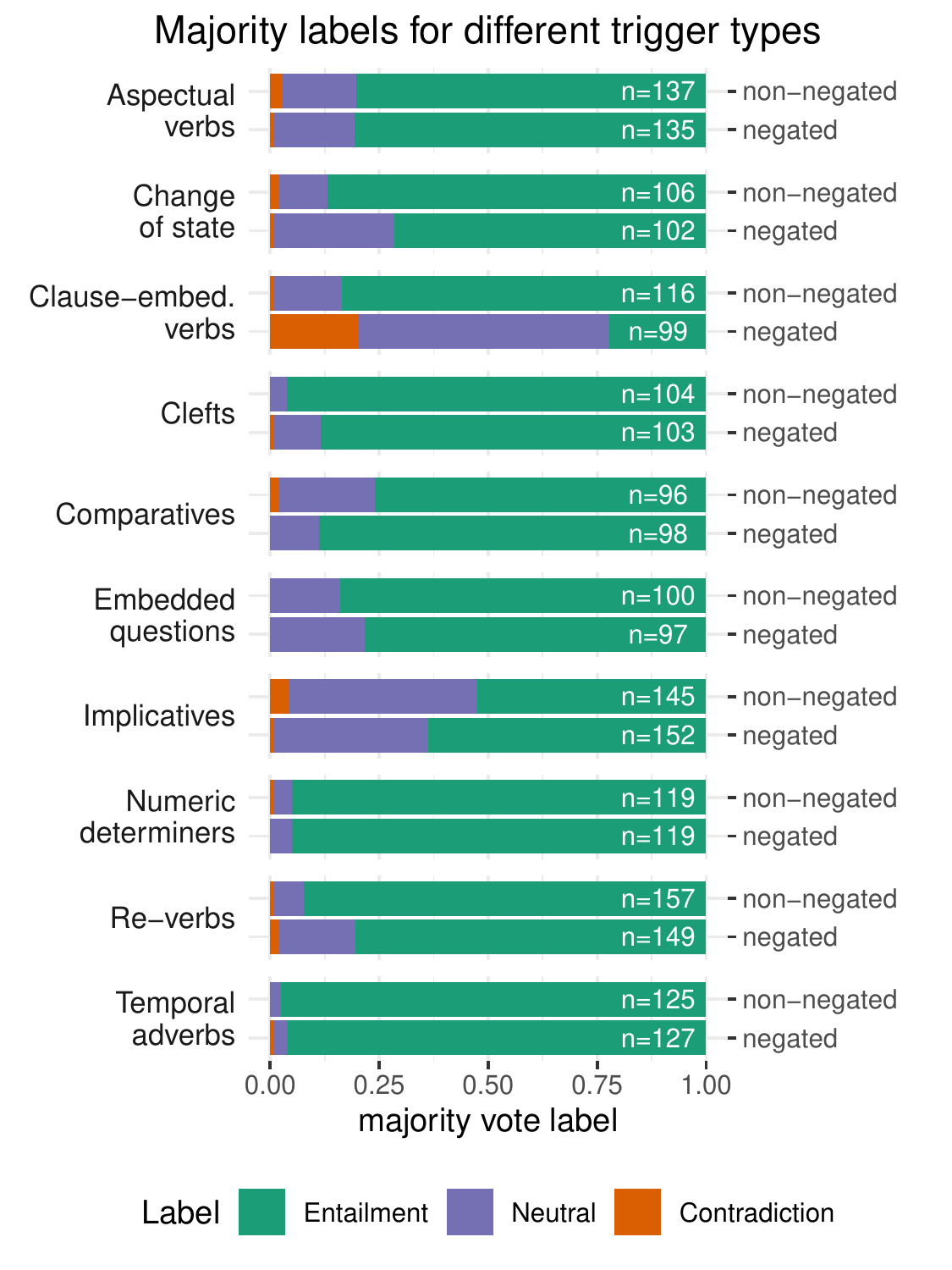}
    \caption{Distributions of mapped human NLI labels for examples in the main corpus, aggregated by trigger type. $n$ indicates the number of examples.}
    \label{fig:aggregate_labels}
\end{figure}

Figure~\ref{fig:aggregate_labels} shows the distribution of labels for each type of trigger in the main corpus, for non-negated and negated premises. In cases in which the premise presupposes the hypothesis, we expect the label for the non-negated and the negated example to be \textsc{entailment} (the entailment-canceling negation should not affect the presupposition); in cases in which the hypothesis is not presupposed, we expect that it is either not entailed by the negated premise or not entailed by both types of premises.

Across all triggers, most examples were labeled as \textsc{entailment}, with the remaining examples largely labeled as \textsc{neutral}.\footnote{Note that this pattern cannot be attributed to a response bias of generally assigning higher probabilities since we found that participants used the entire scale on the filler items.} This pattern suggests that participants judged most of our trigger sentences as presupposing the hypothesis, and their judgments were invariant to the negation of the premise. One exception to this finding was clause-embedding verbs: While the non-negated premises (e.g., \textit{Airline officials, flight attendants and frequent travelers \textbf{say} \ul{Brown's case is hardly unique}}) were often judged as entailing the hypothesis (underlined), this is no longer the case for the negated premises (e.g., \textit{Airline officials, flight attendants and frequent travelers \textbf{don't say} Brown's case is hardly unique}). 
This result is expected, for two reasons. First, our clause-embedding verbs included non-factive verbs: although we generally do not expect a clause embedded under \textit{think} or \textit{say} to be presupposed, we included these verbs as there is no clear-cut distinction between factive and non-factive verbs \citep{tonhauser2020which}. Second, there can be significant variability across contexts, even for verbs like \textit{know} that are commonly assumed to be factive \citep{demarneffe2019commitmentbank}.

The second exception to the general pattern was implicatives, which have a higher proportion of \textsc{neutral} labels than other triggers. This difference reflects the fact that, unlike what is generally assumed in the implicatives literature \cite[e.g.,][]{karttunen1971implicative}, statements such as \textit{X took effort} which are supposedly always triggered by \textit{managed to X} did not always receive full endorsement by participants and were therefore often mapped to neutral.

\begin{table}\small\centering
\newcommand\rotation{20} 
\arrayrulecolor{white} 
\renewcommand{\arraystretch}{1.1}
\begin{tabular}{l|E|E|E|E} 
\multicolumn{1}{p{2.5ex}}{\rotatebox{0}{\textbf{Trigger}}} &
\multicolumn{1}{p{2.5ex}}{\rotatebox{\rotation}{No change}} &
\multicolumn{1}{p{2.5ex}}{\rotatebox{\rotation}{E $\rightarrow$ \{N,C\}}} &
\multicolumn{1}{p{2.5ex}}{\rotatebox{\rotation}{\{N,C\} $\rightarrow$ E}} &
\multicolumn{1}{p{2.5ex}}{\rotatebox{\rotation}{Other}} \\ 
Aspectual verbs & 74.8 & 11.5 & 13.0 & 0.8 \\ \hline 
Change of state & 73.0 & 21.0 & 6.0 & 0.0 \\ \hline 
Clause-embed. verbs & 31.3 & 62.7 & 3.0 & 3.0 \\  \hline 
Clefts & 88.3 & 9.7 & 1.9 & 0.0 \\ \hline 
Comparatives & 76.1 & 5.4 & 17.4 & 1.1 \\ \hline  
Embedded questions & 82.1 & 12.6 & 5.3 & 0.0 \\  \hline 
Implicatives & 70.0 & 7.1 & 21.4 & 1.4 \\ \hline 
Numeric determiners & 95.0 & 2.5 & 1.7 & 0.8 \\ \hline 
Re-verbs & 78.6 & 16.6 & 4.8 & 0.0 \\ \hline 
Temporal adverbs & 96.0 & 2.4 & 1.6 & 0.0 \\ 
\end{tabular}
\caption{Percent of times the label assigned via majority vote of participants to a given example changed when the premise was negated. E=\nli{entailment}, N=\nli{neutral}, and C=\nli{contradiction}.}
\label{tab:label_switch}
\end{table}
\arrayrulecolor{black} 

\paragraph{Context-sensitivity}
The high rates of \nli{entailment} labels for both non-negated and negated examples confirm the standard view that presuppositions project out of negation. More surprising are cases where one or both hypotheses are not entailed. In one sub-case, a presupposition of the non-negated sentence is canceled by negation. We observe this when a non-negated sentence is labeled \nli{entailment}, and the negated sentence either \nli{neutral} or \nli{contradiction} (E $\rightarrow$ \{N,C\} in Table~\ref{tab:label_switch}). Generally, only presuppositions of soft triggers can be canceled under certain contextual conditions \cite{abusch2002lexical,simons2010projects}.

As expected, presupposition cancellation is extremely uncommon for hard triggers (\trigger{numeric determiners} and \trigger{clefts}) which should always give rise to a presupposition, and high for \trigger{clause-embedding verbs} which do not all presuppose their clausal complement. There was also a non-negligible number of examples with other triggers such as \trigger{change of state} verbs and \trigger{re-verbs} whose purported presupposition does not survive negation, which indicates a moderate level of context-sensitivity.
\footnote{For example, given the premise \textit{Brother S (didn't) burst into tears}, participants judge that the hypothesis \textit{Brother S wasn't crying before} holds, but judgments most often map on to \nli{neutral} with the negated premise. This may be due to the use of \textit{burst} with \textit{into tears} since one can softly cry and then burst into tears, which makes the presupposition weaker in this context.}

\begin{figure}
    \centering
    \includegraphics[trim=0 0.2cm 0 0,clip,width=0.9\linewidth]{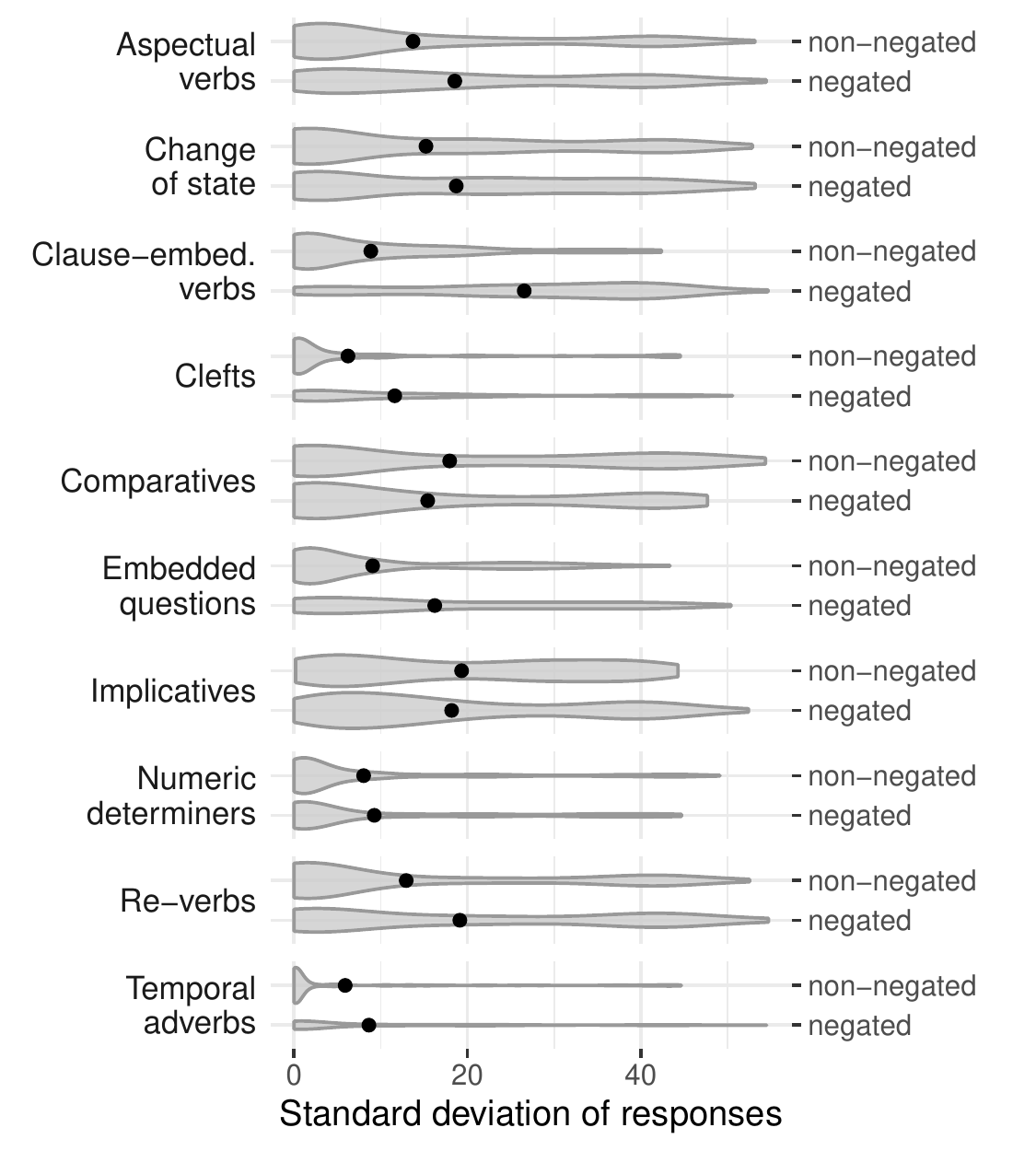}
    \caption{Distributions of standard deviations of responses for each example across all trigger types. Dots represent the mean standard deviation.}
    \label{fig:stdev}
\end{figure}

Examples with \trigger{implicatives}, \trigger{comparatives}, and \trigger{aspectual verbs}, however, sometimes exhibited the opposite behavior: the non-negated example was labeled \nli{neutral} or \nli{contradiction} whereas the negated example was labeled \nli{entailment}. Through manual inspection of these examples, we found that these cases involve a range of semantic and pragmatic inferences.\footnote{For example, for the non-negated version of the trigger sentence with a comparative \textit{The PCA results suggest that ecosystem type is a stronger predictor of the index profiles than sequence quality}, participants judged that it does not entail that \textit{sequence quality is a predictor}, presumably because participants concluded that a weaker predictor is potentially no predictor at all.}

\paragraph{Gradience}

To determine the extent of gradient judgments, i.e., judgments that don't lie at either end of the probability scale, we considered both the proportion of \nli{neutral} labels  (see Figure~\ref{fig:aggregate_labels}) as well as the variability in the continuous ratings for each example. A high proportion of \nli{neutral} labels indicates that participants provided many ratings in the middle of the scale, and likewise, higher variability indicates not just that humans disagree more but also that they use a larger proportion of the scale as opposed to just the endpoints.

As we already mentioned, the proportion of \nli{neutral} labels was particularly high for \trigger{implicatives} and \trigger{clause-embedding verbs}, and greater than 10\% for all trigger types other than \trigger{temporal adverbs} and  the hard triggers \trigger{clefts} and \trigger{numeric determiners}.

Figure \ref{fig:stdev} shows the distributions of the standard deviation across all five probability judgments for each example. These distributions confirm the findings from analyzing the proportion of \nli{neutral} labels: Variability within examples was highest for \trigger{clause-embedding verbs} and \trigger{factives}; and lowest for \trigger{temporal adverbs}, \trigger{clefts} and \trigger{numeric determiners}. Taken together, these patterns provide evidence that there exists at least a moderate amount of gradience for all triggers except for the hard triggers and temporal adverbs.\footnote{For certain triggers, some \nli{neutral} ratings could reflect inconsistencies in stating the presupposition in English, rather than inherent gradience. A reasonable, semantically stable presupposition might exist in a logical metalanguage. This caveat applies to \trigger{change of state} predicates and \trigger{implicatives} where the presupposition statement cannot be mechanically derived from an expression in the trigger sentences.}

\subsection{Adversarial Examples}
\begin{figure}
    \centering
    \includegraphics[trim=0 0.4cm 0.5cm 0,clip,width=0.9\linewidth]{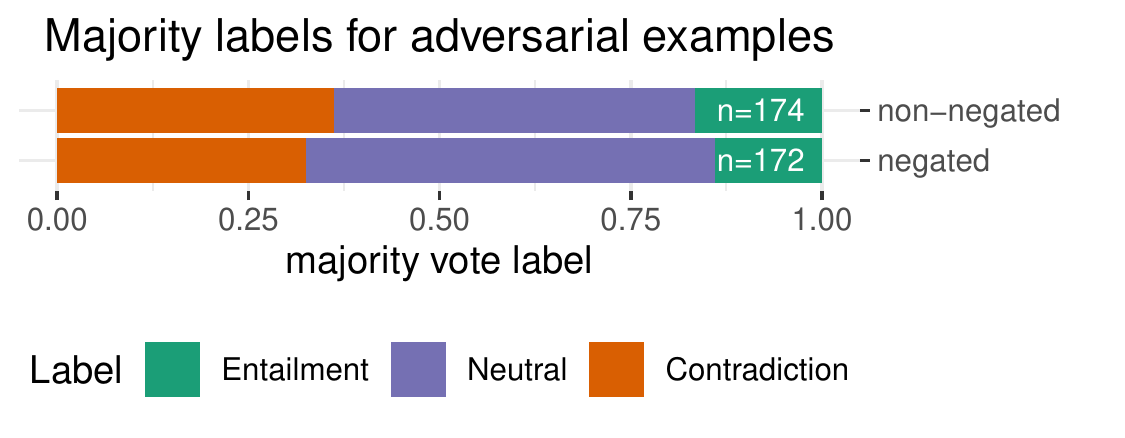}
    \caption{Distributions of mapped human NLI labels for examples in the adversarial corpus. $n$ indicates the number of examples.}
    \label{fig:adversarial_labels}
\end{figure}

Figure~\ref{fig:adversarial_labels} shows the distribution of labels for the adversarial examples. As intended with these additional examples, the hypotheses in most of the adversarial examples were judged to be no longer entailed by the premise. The small proportion of examples labeled \textsc{entailment} can largely be attributed to noise from the label-mapping procedure as well as participants being fooled by very subtle changes to the hypothesis (e.g., assigning a rating of 100 to an example in which \textit{John Doe and Jim Miller} was replaced with \textit{Jim Doe and John Miller}).

\subsection{Judgment Consistency}

To determine whether the human judgments were sufficiently consistent for evaluating models, we computed several agreement statistics commonly used for NLI datasets, and compared them to the statistics for existing NLI corpora.  Table~\ref{tab:accuracy_comparison} shows how often the (mapped) labels of all participants agree, as well as how often individual (mapped) labels agree with the majority label. Despite the potentially noisy mapping from continuous ratings to categorical labels, the agreement numbers are comparable to the ANLI corpus which, like NOPE, contains many challenging examples.

\begin{table}[]
\centering
\resizebox{\linewidth}{!}{%
    \begin{tabular}{lllll}
    \toprule
    \textbf{Measure} & \textbf{SNLI} & \textbf{MNLI} & \textbf{ANLI} & \textbf{NOPE} \\
    \midrule
    Unanimous agreement & 58.3 & 58.2 & 41.6 & 38.7 \\
    Individual lab. = majority lab. & 89.0 & 88.7 & 81.7 & 81.5 \\
    No majority & 2.0 & 1.8 & 13.0 & 5.1 \\
    \bottomrule
    \end{tabular}
}
\caption{Validation statistics (\%) reported for SNLI, MNLI, and ANLI compared with NOPE.}
\label{tab:accuracy_comparison}
\end{table}

\section{Machine Learning Experiments} 

\subsection{Models}
\label{sec:models}

To evaluate the ability of NLI models to infer presuppositions, we evaluated two baseline models, a Bag-of-Words (BOW) model and InferSent \cite{conneau-etal-2017-supervised}, as well as two pre-trained  transformer models, RoBERTa-large \cite{liu2019roberta} and DeBERTa-V2-XLarge \cite{he2020deberta}, which both recently achieved state-of-the-art performance on the SuperGLUE natural language understanding benchmark \cite{wang2019superglue}. 

The BOW model produces a sentence representation from the mean of FastText word vectors \cite{mikolov2018advances}, and InferSent does so using a bidirectional LSTM. We trained baselines end-to-end on the combination of MNLI \cite{williams2020mnli}, SNLI \citep{bowman2015snli}, ANLI \citep{nie2020adversarial}, and FEVER \citep{thorne2018fever}. For each baseline model type and training set, we trained 16 models, and evaluated the 5 with highest validation accuracy. We adapted \citeauthor{conneau-etal-2017-supervised}'s code to train and evaluate these baselines, and include this code in the NOPE codebase.

RoBERTa-large and DeBERTa-V2-XLarge are both transformer-based masked language models with 355M and 900M parameters, respectively. We fine-tuned these models on the above combination of NLI datasets using the HuggingFace Transformers library \cite{wolf2020transformers} through an adapted version of the scripts by \citet{nie2020adversarial}. We fine-tuned RoBERTa five times with different random seeds. Due to the high computational cost of fine-tuning DeBERTa, we only fine-tuned it once.

\subsection{Results and Discussion}

\begin{figure*}
    \centering
    \includegraphics[trim=0 0 0 0.1cm,clip,width=\textwidth]{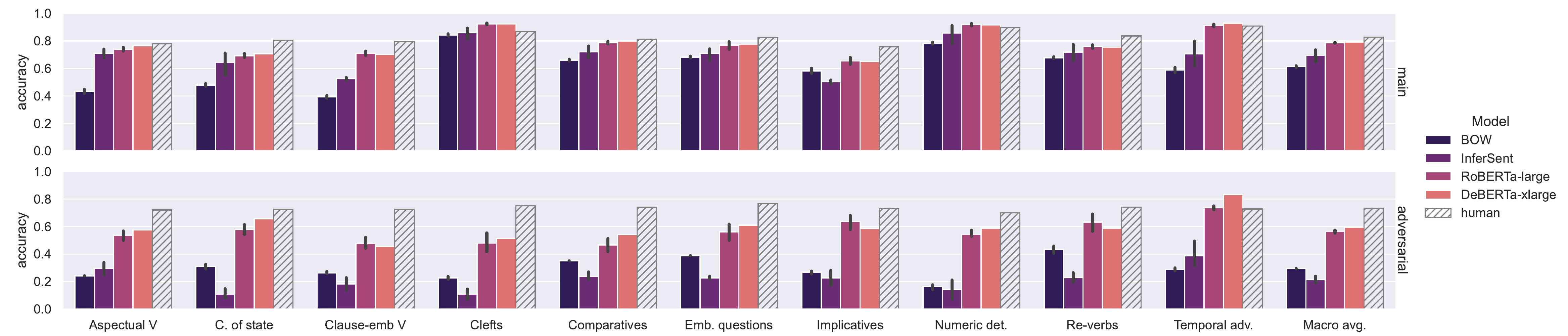}
    \caption{Accuracy on NOPE for both the main and adversarial versions of the corpus. Human scores correspond to the proportion of responses that agreed with the majority label.}
    \label{fig:nope_acc}
\end{figure*}

As Figure~\ref{fig:nope_acc} shows, the transformer models achieve high accuracy on the main dataset, and in particular for the trigger types clefts, numeric determiners, and temporal adverbs, performance is almost at ceiling. The BOW and the InferSent baselines achieve much lower overall accuracy, highlighting that NOPE is a challenging evaluation dataset.

\paragraph{Lexical overlap bias} All models achieve lower accuracy on the adversarial data than on the main data, suggesting that all models exhibit a bias for predicting \nli{entailment} when lexical overlap is high. This is particularly true for the two baseline models, which suggests that a large amount of their success on the main dataset could be driven by the lexical overlap between the spelled-out presupposition and the trigger sentence rather than proper inferences involving presuppositions. We thus exclude the baseline models from the subsequent analyses.
The performance gap between the main corpus and the adversarial corpus for transformer models, however, is much smaller, which suggests that these models do not primarily rely on a lexical overlap heuristic when making predictions.

\begin{table}[]
    \centering\footnotesize
\renewcommand{\arraystretch}{1.1}
    \begin{tabular}{l F F F F F F}
    \textbf{Model} & \multicolumn{2}{c}{E \tikzmark{x1} \phantom{xxxx} \tikzmark{x2} E} & \multicolumn{2}{c}{E \tikzmark{x3} \phantom{xx} \tikzmark{x4} \{N,C\}} & \multicolumn{2}{c}{\{N,C\} \tikzmark{x5} \phantom{xx} \tikzmark{x6} E} \\
    {} & \multicolumn{1}{p{2.5ex}}{nonneg} &
    \multicolumn{1}{p{2.5ex}}{\phantom{x}neg} & \multicolumn{1}{p{2.5ex}}{nonneg} & \multicolumn{1}{p{2.5ex}}{\phantom{}neg} & \multicolumn{1}{p{2.5ex}}{nonneg} & \multicolumn{1}{p{2.5ex}}{\phantom{xx}neg} \\
    \midrule
    {\scriptsize RoBERTa} & 89.9 & 89.2 & 80.6 & 32.7 & 35.1 & 68.4 \\
    {\scriptsize DeBERTa} & 90.8 & 88.6 & 81.8 & 32.1 & 38.9 & 68.9 \\
    \end{tabular}
    \caption{Model accuracy on non-negated and negated examples for different subsets of the main corpus. ``E~$\rightarrow$ E'' corresponds to examples in which the presupposition projects out negation and ``E $\rightarrow$ \{N,C\}'' and ``\{N,C\} $\rightarrow$ E'' correspond to examples in which negation affects whether the hypothesis is entailed.}
    \label{tab:detailed_model_res}
    \begin{tikzpicture}[remember picture,overlay]
    \draw[-latex] ([shift={(0ex,0.5ex)}]pic cs:x1) -- ([shift={(0ex,0.5ex)}]pic cs:x2);
    \draw[-latex] ([shift={(0ex,0.5ex)}]pic cs:x3) -- ([shift={(0ex,0.5ex)}]pic cs:x4);
    \draw[-latex] ([shift={(0ex,0.5ex)}]pic cs:x5) -- ([shift={(0ex,0.5ex)}]pic cs:x6);
    \end{tikzpicture}
\end{table}

\paragraph{Projection} To investigate whether models draw correct inferences in cases where the presupposition projects out of negation, i.e., whether they predict \nli{entailment} for both the non-negated and the negated version of the trigger sentence, we computed model accuracy on the subset of the data for which both the non-negated and negated versions of the same example were labeled \nli{entailment} (``E $\rightarrow$ E'' in Table~\ref{tab:detailed_model_res}). Accuracy was high (>88\% for both models) for both non-negated and negated sentences, suggesting that models often draw the correct inferences for sentences in which the presupposition projects out of negation.

\paragraph{Context-sensitivity} To investigate the models' behavior on examples in which the trigger sentence does not give rise to a presupposition, we also computed the accuracy on the subset of the main corpus in which the mapped human label differed between the negated and non-negated trigger sentence (columns ``E $\rightarrow$ \{N,C\}'' and ``\{N,C\} $\rightarrow$ E'' in Table~\ref{tab:detailed_model_res}). These correspond to sentences without actual presupposition triggers (e.g., non-factives such as \textit{think}) and sentences in which the context cancels the presupposition. Accuracy on this subset of the data is much lower, especially for sentences in which the hypothesis is not entailed, suggesting that the models do not exhibit the same sensitivity to negation in the trigger sentences as humans. This could either be because the models are not sensitive to negation at all, as found for other models \cite{ettinger2020bert}, or fail to draw the correct inferences when the context cancels a presupposition.

\paragraph{Gradience} Finally, to investigate the models' ability to capture the gradience in human judgements, we computed the model accuracy on the subset of examples that were assigned a \nli{neutral} label. Accuracy on these examples is again much lower (39.2\% for RoBERTa, and 39.1\% for DeBERTa) than accuracy on the overall corpus, which suggests that the transformer models do not draw the correct inferences for trigger sentences for which  participants neither fully endorsed nor rejected the hypothesis.

\section{Conclusion}

We presented NOPE, a dataset of naturally occurring presuppositions involving a diverse set of presupposition triggers. The human judgments in this dataset suggest that the examples we identified give rise to presuppositions in most cases, and that presupposition cancellation and variability exist in a small but non-negligible minority of cases across contexts for presuppositions other than those triggered by clause-embedding verbs and implicatives. We used this dataset to evaluate the pre-trained transformer-based models RoBERTa and DeBERTa. We found that the models were able to draw inferences involving presuppositions in the simple cases but failed to fully capture human-level context-sensitivity and gradience.

\section*{Acknowledgements}
We thank the three anonymous reviewers and the meta reviewer for their thoughtful feedback. This project has benefited from financial support to SB by Eric and Wendy Schmidt (made by recommendation of the Schmidt Futures program), Samsung Research (under the project \textit{Improving Deep Learning using Latent Structure}), Apple, and Intuit, and from in-kind support by the NYU High-Performance Computing Center. SS was in part supported by the Stanford HAI Hoffman-Yee Project \textit{Towards Grounded, Adaptive Communication Agents}. This material is based upon work supported by the National Science Foundation under Grant No. 1850208, as well as under Grant No. 2030859 to the Computing Research Association for the CIFellows Project. Any opinions, findings, and conclusions or recommendations expressed in this material are those of the
authors and do not necessarily reflect the views of the National Science Foundation nor the Computing
Research Association.

\bibliography{anthology,acl2020}
\bibliographystyle{acl_natbib}

\newpage
\null
\newpage

\appendix

\section{Trigger Types}
\label{sec:trigger_types}

\paragraph{Change of state verbs}
Change of state verbs such as \textit{appear} and \textit{snap} presuppose that the entity that was affected by the described event was in a different state just before the event happened. For instance, the sentence \textit{Cats appeared on the street} presupposes that \presup{cats had not been on the street right before then}.

\paragraph{Clefts}
Cleft constructions such as \textit{It was my cat that made a noise} take the form \textit{It was X who/that did Y} This example presupposes that \presup{something made a noise}, i.e., \presup{someone/something did Y} (see \citealt{delin1992properties, delin1995presupposition, prince1986syntactic, soames1982presuppositions}). Note that this example also carries the additional inference that \presup{my cat is the \textbf{only} thing that made a noise}, but to keep matters simple, we consider only one presupposition per trigger type and therefore ignore this second inference for present purposes. 

\paragraph{Comparatives}
Comparative constructions such as \textit{Sandy is a bigger cat than Holly} take the form \textit{X is a W-er Y than Z}. The example provided above presupposes that \presup{Holly is a cat} (i.e., \presup{Z is a Y}) (see \citealt{levinson1983pragmatics,cummins2012experimental,cummins2013backgrounding}).

\paragraph{Aspectual verbs}
Aspectual verbs such as \textit{start} and \textit{stop} presuppose whether the event that is embedded under these verbs had previously been happening or not (see \citealt{abrusan2011predicting, abusch2002lexical}). The sentence \textit{Lisa stopped petting Tom's cat} presupposes that \presup{Lisa had previously been petting Tom's cat}. Aspectual verbs are frequently subsumed under change of state verbs but unlike the verbs we consider change of state verbs, aspectual verbs take a non-finite verb phrase (e.g., \textit{petting Tom's cat} or \textit{to eat the kibble}) as a complement.

\paragraph{Embedded questions}
Embedded questions are realized when a clause is embedded under a wh-word such as \textit{why}, \textit{how}, \textit{where}, or \textit{when} as in \textit{Julia knows why Lisa likes Tom's cat} (see \citealt{karttunen1977syntax, groenendijk1991dynamic, uegaki2019semantics}). 
These triggers presuppose the truth of the embedded content; the example provided above presupposes that \presup{Lisa likes Tom's cat}.

\paragraph{Clause-embedding predicates}
Clause-em\-bed\-ding predicates frequently presuppose the truth of the finite clause that they embed. This set includes verbs such as \textit{realize}, \textit{know}, and \textit{regret} \cite{abusch2004empty, beaver2001presupposition, heim1992presupposition, karttunen1973presuppositions, karttunen1974presupposition}. The sentence \textit{Julia regrets having forgotten to feed Holly} presupposes that \presup{it is true that Julia forgot to feed Holly}.

Importantly, not all clause-embedding predicates presuppose their complement. For example, predicates such as \textit{think} or \textit{say} do not necessarily entail their clausal complement. However, unlike originally assumed there exists no clear-cut distinction between clause-embedding predicates that do and ones that do not presuppose their complement \cite[e.g.,][]{demarneffe2019commitmentbank, tonhauser2020which}, and therefore we include all clause-embedding predicates that appeared in COCA in the dataset presented in this paper.

\paragraph{Implicative predicates}
Implicative verbs such as \textit{manage to} and \textit{fail to} presuppose some property of the action in the clause that they embed (see \citealt{karttunen1971implicative, karttunen1979conventional}). For example, the sentence \textit{Holly failed to escape her pet taxi} presupposes that \presup{Holly attempted to escape her pet taxi}, since \textit{fail to} implies an attempt at the action. Likewise, \textit{Holly managed to X} presupposes that \presup{X would take effort for Holly}.

\paragraph{Numeric determiners}
Numeric determiners such as \textit{both} or \textit{all X}, where X is numeric expression such as \textit{three} presuppose that there is a precise number of modified entities in the context (see \citealt{lappin1988presuppositional}). The sentence \textit{All three cat owners that Julia spoke to want another cat} presupposes that \presup{there are three cat owners that Julia spoke to}. We extracted sentences that contain \textit{both (of the) N}, \textit{all NUM (of the) N}, or \textit{all (of the) NUM N}, where \textit{N} is a noun with optional modifiers, and \textit{NUM} is a numeric expression.

\paragraph{\textit{Re-}prefixed verbs}
Verbs with the prefix \textit{re-} presuppose that the action of the verb (attaching to \textit{re-}) had taken place in the past (see \citealt{beaver1997presupposition, marantz2009roots, wechsler1989accomplishments}). The sentence \textit{Holly re-entered the room} presupposes that \presup{Holly had entered the room before}. Hence, \textit{re-V} presupposes that \presup{V had been carried out before}.  

\paragraph{Temporal adverbs}
Adverbial embedded clauses headed by prepositions such as \textit{before}, \textit{after}, \textit{since}, and \textit{while} presuppose the content of the clause they embed (see \citealt{beaver2003uniform, heinamaki1974semantics}). The sentence \textit{Lisa petted Tom's cat after she washed her hands} presupposes that \presup{Lisa washed her hands}. We extracted sentences with adverbial clauses headed by \textit{after}, \textit{since}, \textit{before}, \textit{because}, and \textit{while}.

\section{Word Lists}
\label{sec:word-lists}

Many of the presupposition triggers included in NOPE depend on certain lexical items. We used the following word lists for extracting examples with lexical triggers.

\paragraph{Change of state verbs} We compiled a list of common change of state verbs by manually extracting unambiguous change of state verbs from the 250 most frequent verbs in COCA, resulting in the following list of 43 verbs: \textit{appear}, \textit{arrive}, \textit{ascend}, \textit{break}, \textit{burst}, \textit{clean}, \textit{close}, \textit{collapse}, \textit{crack}, \textit{crash}, \textit{curl}, \textit{descend}, \textit{die}, \textit{drop}, \textit{enter}, \textit{erupt}, \textit{escape}, \textit{explode}, \textit{expose}, \textit{fall}, \textit{fill}, \textit{fire}, \textit{fix}, \textit{freeze}, \textit{graduate}, \textit{hide}, \textit{hire}, \textit{leave}, \textit{lose}, \textit{melt}, \textit{open}, \textit{pop}, \textit{remain}, \textit{return}, \textit{rise}, \textit{shut}, \textit{sink}, \textit{snap}, \textit{split}, \textit{stay}, \textit{tear}, \textit{wake}, \textit{win}.

\paragraph{Aspectual verbs} We used the verbs from the Levin verb classes \cite{levin1993english} 55.1 (``begin'' verbs) and 55.2 (``complete'' verbs), resulting in the following list: \textit{begin}, \textit{cease}, \textit{commence}, \textit{complete}, \textit{continue}, \textit{discontinue}, \textit{end}, \textit{finish}, \textit{halt}, \textit{initiate}, \textit{keep}, \textit{proceed}, \textit{quit}, \textit{repeat}, \textit{resume}, \textit{start}, \textit{stop}, \textit{terminate}.

\paragraph{Embedded questions} We used the following wh-words (in combination with syntactic patterns) to extract sentences with embedded questions:
\textit{why}, \textit{how}, \textit{where}, \textit{when}, \textit{who}, \textit{what}, \textit{which}.

\paragraph{Implicative predicates} We used the following list of implicative predicates, adapted from the list by \citet{karttunen1971implicative}: \textit{avoid}, \textit{bother}, \textit{care to}, \textit{condescend to}, \textit{dare}, \textit{decline}, \textit{fail}, \textit{forget to}, \textit{happen to}, \textit{have the misfortune}, \textit{manage to}, \textit{neglect}, \textit{refrain}, \textit{remember to}, \textit{resist}, \textit{see fit}, \textit{take the time}, \textit{take the trouble}, \textit{venture}.

\section{Writing Guidelines} 
\label{sec:writing-guidelines}

In constructing the presupposition statement (the hypothesis) for each example, we adhered to the following  guidelines:
\begin{itemize}[noitemsep]
    \item If a first person pronoun is used in the premise, use a first person in the hypothesis
    \item If the premise uses a third person pronoun, replace it with the full antecedent from context, even if the full antecedent is in a sentence from the prior context rather than the sentence with the trigger.
    \item If the presupposition trigger is inside a quotation, it can be included only if it is part of a simple speech tag. In this case, use the same pronoun information from the premise in the hypothesis
    \item If the presupposition trigger is already embedded under negation, create a non-negated minimal pair premise. 
    \item Make small edits to the premise if it makes an otherwise unviable example able to be included. If edits are made, make them to both the negated and non-negated premises. Examples of common edits include:
    \begin{itemize}[noitemsep]
        \item Removing an entire conjunct
        \item Changing \textit{but} or \textit{or} to \textit{and} (and vice versa)
        \item Removing additional words like \textit{even}, \textit{so}, or \textit{also} that can be unnatural with negation
        \item Removing NPIs like \textit{yet} that would not be licensed when the negation is removed
    \end{itemize}
    \item When there are multiple conjuncts that contain negateable verbs, only negate the conjunct that contains the trigger.
\end{itemize}

\paragraph{Trigger-specific guidelines}
Since the presupposition that arises from different triggers can be very specific to that trigger, we also followed some trigger-specific guidelines:
\begin{itemize}[noitemsep]
    \item \trigger{Aspectual verbs}: The reference time is important for judging the presupposition, so adjust the tense of the presupposition to match a more natural-sounding phrase rather than keeping the exact tense from the trigger sentence.
    \item \trigger{Change of state predicates}: Adjust tense or aspect as needed when referring to the past to make sure the sentence sounds natural, while the presupposition still holds. Can adjust tense/aspect in ways that do not affect truth conditions.
    \item \trigger{Clause-embedding verbs}:  No trigger specific guidelines.
    \item \trigger{Clefts}: No trigger specific guidelines.
    \item \trigger{Comparatives}: No trigger specific guidelines.
    \item \trigger{Embedded questions}: No trigger specific guidelines. 
    \item \trigger{Implicative predicates}: For \textit{manage}, make sure the wording is something like \textit{it \textbf{would} take effort to X} instead of \textit{it \textbf{took}/\textbf{takes} effort to X} to allow the presupposition to hold under negation. For example, if the premise is \textit{They managed to make it to shore}, then the presupposition \textit{it took effort to make it to shore} would no longer hold with the negated premise \textit{They didn't manage to make it to shore}, but this is not due to the presupposition failing to project, as a presupposition phrased as \textit{it would take effort to make it to shore} holds in both the negated and non-negated premises.
    \item \trigger{Numeric determiners}: Edit the original (and negated) premise to include a specific domain restriction, and reference that domain restriction in the hypothesis. For example, \textit{both men talked} can become \textit{both men in the room talked} to allow the presupposition to be \textit{there are exactly two men in the room} as opposed to the less natural \textit{there are exactly two men}. We found that statements like \textit{there are exactly two men} were rated as very improbable by participants, even though the statement holds of the event in the premise, but adding a domain restriction that did not otherwise affect the truth conditions of the sentence allowed for a more natural presupposition. When the domain restriction is unstated but highly salient (e.g., \textit{both hands}), it does not need to be added to the premise.
    \item \trigger{Re-verbs}: Adjust tense/aspect as needed to keep the presupposition sounding natural, so long as the changes do not affect truth conditions. Where possible, use the passive form of the presupposition, as the presuppositions with \textit{re-} hold for the event but not necessarily the event agent.
    \item \trigger{Temporal adverbs}: If the adverb phrase begins the sentence (e.g., \textit{after Jody left, I called Bill}), reverse the order of clauses so that when the matrix clause is negated, the negation precedes the temporal adverb.
\end{itemize}

\paragraph{Metadata recorded}
We additionally recorded many properties of the example sentences that may have an effect on the strength of the presupposition. 
Though full analysis of these components is beyond the scope of this paper, the full corpus metadata includes expert annotations indicating if any of the following hold:
\begin{itemize}[noitemsep]
    \item \textit{Small edits}: We indicate any cases where we made small edits (e.g., changing `but' to `and') to the original sentence that was pulled from COCA.
    \item \textit{Conjunction}: We indicate whether the premise contains conjunction, regardless of whether the conjoined phrases are DPs, VPs, or CPs.
    \item \textit{Original is negated}: We indicate if, in the original premise, the presupposition trigger is already embedded under negation, including if the matrix verb is negated, there is a negative existential, and if there's a negative quantifier.
    \item \textit{Innocent embedding}: We indicate cases where the trigger is embedded, but in a way that is not complicated or does not affect the interpretation of the presupposition (e.g. in a quotation embedded under a speech tag).
\end{itemize}

\section{Crowdsourcing Experiment}
\label{sec:crowdsourcing-appendix}

As mentioned in the main text, we used a pre-screener to find high quality participants, and we excluded items from participants who performed poorly on filler items, as described in the following paragraphs.

\paragraph{Pre-screener}

The pre-screener was open to participants in the US with approval ratings greater than 98\% and more than 10,000 previously-completed HITs. It included the same task as the main rating task with 10 hand-selected items to test basic numerical reasoning and the use of the full rating scale. To ensure that participants understood how the slider worked, they completed three practice trials that provided feedback on their use of the slider before they continued to the 10 screening trials.

For each item in the screener, we determined the acceptable range of ratings in advance (e.g., for a premise such as \textit{Jody loves dogs so she went out and adopted a cute little poodle}, the range of acceptable judgments for the hypothesis \textit{Jody adopted a dog} would be anywhere from 95 to 100).
To qualify for the main task, participants had to achieve an accuracy of 70\% on the pre-sceener.
A total of 150 participants completed the screener, of which 107 scored above the accuracy threshold and were given the qualification to complete the main task.

\paragraph{Exclusions}

In the main task, we used the filler items to filter participants who did not seem to pay attention. Considering that we took the items from the existing MNLI dataset, we knew the entailment relation between the premise and hypothesis. We could therefore compare whether participants provided ratings consistent with the entailment relation or not and use these comparisons as attention checks. Specifically, we counted ratings on fillers with an \textsc{entailment} label as correct if a participant provided a rating above 90; and we counted ratings on fillers with a \textsc{contradiction} label as correct if a participant provided a rating below 10.\footnote{These cutoffs are loosely based on the average ratings for \textsc{entailment} and \textsc{contradiction} on the portion of SNLI that \citet{chen2020uncertain} re-annotated with probability ratings.} We excluded data from in total 15 participants whose accuracy on fillers was below 70\%, and collected additional ratings such that we obtained 5 ratings from high-accuracy participants for each item.  

\section{NLI Label Mapping Procedure}
\label{sec:nli-label-mapping}

As mentioned in the main text, for each participant, we inferred an upper threshold for mapping a probability rating to the NLI label \textsc{contradiction} and a lower threshold for \textsc{entailment}. Probability ratings that fall in between these two thresholds were mapped to \textsc{neutral}. To infer the lower threshold for \textsc{entailment}, we considered the filler examples whose labels in the MNLI dataset are either \textsc{neutral} or \textsc{entailment}. Further, since we expected the threshold to lie above 50\%, we only considered examples with a probability rating greater than 50\% to filter out noise. We created participant-specific datasets for each participant by upweighting the participant's ratings by a factor of 50. This combination of upweighted participant-specific ratings and global ratings was intended to infer thresholds that better reflect the participant's use of the scale while at the same time avoiding extreme thresholds through still considering the ratings by all other participants. We then optimized the threshold such that the accuracy of the mapping from probability ratings to NLI labels is maximized for the weighted examples under consideration. 

To infer the upper threshold for \textsc{contradiction}, we applied the same procedure to the filler examples whose labels are either \textsc{contradiction} or \textsc{neutral} and which received a rating below 50\%.

\section{Comparison to ImpPres}

\citet{jeretic2020impppressive} automatically generated a dataset (ImpPres) using templates for evaluating NLI models' abilities to draw inferences involving different presupposition triggers. This work made the simplifying assumptions that there exists no contextual variability and therefore that triggers always give rise to presuppositions. 

To compare to what extent the model behavior differs between naturalistic and automatically generated examples, we also evaluated all models  on the ImpPres dataset and compared accuracy for the trigger types that are present in both ImpPres and NOPE.

\begin{figure*}
    \centering
    \includegraphics[width=\textwidth]{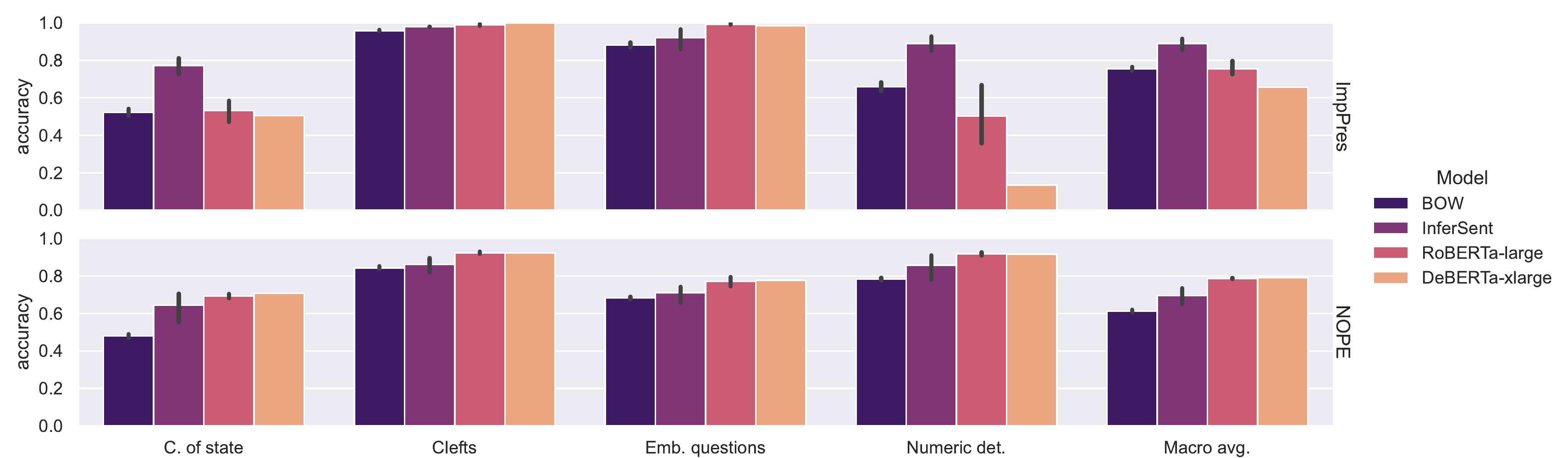}
    \caption{Comparison of model accuracries on ImpPres \cite{jeretic2020impppressive} and NOPE for all trigger types that are present in both datasets.}
    \label{fig:imppres}
\end{figure*}

As Figure~\ref{fig:imppres} shows, accuracy on clefts and embedded question triggers are almost at ceiling on the ImpPres dataset, whereas the same models achieve lower accuracy on the examples in NOPE. In the case of clefts, the difference in accuracy between ImpPres and NOPE is very small and the lower accuracy on clefts on the NOPE dataset might be a result of noise in the human annotations. 

However, the larger gap in the results for embedded questions, for which we found that variability exists in naturally occurring examples, suggests that for this trigger, the models are only able to draw correct inferences when the presupposition is not canceled by the context (which is the case for all examples in ImpPres).

Accuracy on examples with numeric determiners is extremely low for the transformer models evaluated on ImpPres. This seems to be a result of a difference in how the presupposition was spelled out. In ImpPres, all spelled-out presuppositions triggered by numeric determiners include the modifier \textit{exactly} (e.g, \textit{There are \textbf{exactly} three cats on the mat}), whereas we omitted \textit{exactly} from the spelled-out presupposition in NOPE since the presence of \textit{exactly} might trigger additional inferences in humans independent of the presupposition. The low accuracy on ImpPres is caused by the model generally predicting \nli{neutral} when the hypothesis contains \textit{exactly}, which may or may not differ from how humans would judge these examples.

Finally, we also observe differences in accuracy on examples with change of state verbs. Manual inspection of the examples in ImpPres revealed that some of the examples with this trigger have been incorrectly generated (e.g., \textit{Gary did get a job} allegedly entails that \textit{Gary is unemployed}), which explains the lower accuracy on these items in ImpPres and also highlights the importance to verify inferences with human participants.

\section{Model performance on SNLI, MNLI, and ANLI}

We evaluated our NLI models on common NLI validation sets. The results are shown in Table \ref{tab:nli_results}. Whenever possible, we also include previously published results from similar models. Note, however, that with the exception of RoBERTa, these results are not entirely comparable, since our models are trained on a combined set of ANLI, MNLI, SNLI, and FEVER, while other models are generally trained or fine-tuned only on a single dataset. We also report an average over several models, while others report best performance. We generally observe similar performance between our models and previously published ones. The main exception is InferSent, which performs over 10 percentage points worse on MNLI than the model evaluated by \cite{wang-etal-2018-glue}. We found that the best of 16 InferSent models trained using our code on MNLI alone achieved 60.9\% accuracy on both MNLI validation sets.

\begin{table*}[]
    \centering
    \resizebox{\textwidth}{!}{%
    \begin{tabular}{llllllll}
\toprule
& \bf ANLI R1 &  \bf ANLI R2 & \bf ANLI R3 & \bf MNLI-m &\bf MNLI-mm &\bf SNLI &\bf Macro avg \\
\midrule
BOW (ours) & 33.9 (0.81) & 35.6 (0.19) & 33.4 (0.37) & 52.8 (0.33) & 54.2 (0.35) & 59.9 (0.70) &  45.0 (0.30) \\
InferSent (ours) & 34.2 (0.88) & 39.3 (1.37) & 35.7 (1.16) &   54.8 (1.27) & 55.5 (1.03) & 69.8 (1.67) & 48.2 (0.96) \\
RoBERTa-L (ours) & 73.6 (1.07) & 52.3 (0.98) & 49.2 (0.91) & 90.2 (0.17) & 90.0 (0.16) & 92.9 (0.24) & 74.71 (0.16)\\
DeBERTa-XL (ours) & 80 (--) & 62.3 (--) & 60.5 (--) & 91.7 (--) & 91.4 (--) & 93.9 (--) & 80.0 (--) \\\midrule
BOW (\citeauthor{wang-etal-2018-glue}) & -- & -- & -- & 56.0 & 56.4 & -- & -- \\
InferSent (\citeauthor{wang-etal-2018-glue}) & -- & -- & -- & 66.1 & 65.7 & -- & -- \\
RoBERTa (\citeauthor{nie2020adversarial}) & 73.8 & 48.9 & 44.4 & 91.0 & 90.6 & 92.6 & 73.6\\
DeBERTa (\citeauthor{he2020deberta}) & -- & -- & -- & 91.7 & 91.6 & -- & -- \\

\bottomrule
\end{tabular}}
    \caption{Validation set results for our trained NLI models on ANLI, MNLI, and SNLI. Results given are the average percent accuracy (standard deviation) over the top five models (except for DeBERTa). ANLI-R$n$ is the validation set from the $n^\emph{th}$ annotation round. MNLI-m and MNLI-mm are the matched and mismatched subsets of MNLI, respectively.}
    \label{tab:nli_results}
\end{table*}

\end{document}